\RequirePackage[svgnames]{xcolor}
\documentclass[11pt,logo,letterpaper]{berkeley}

\usepackage[comma,authoryear,compress]{natbib}
\usepackage{hyperref}
\IfFileExists{algorithm.sty}{\usepackage{algorithm}}{\usepackage{float}\newfloat{algorithm}{tbp}{loa}\floatname{algorithm}{Algorithm}}
\IfFileExists{algpseudocode.sty}{\usepackage{algpseudocode}}{\newcounter{algorithmicline}
\newenvironment{algorithmic}[1][]{%
  \setcounter{algorithmicline}{0}%
  \par\begingroup\setlength{\parindent}{0pt}%
  \def\algorithmicindent{1.5em}%
  \def\State{\par\stepcounter{algorithmicline}\llap{\scriptsize\thealgorithmicline\quad}}%
  \def\Statex{\par\hspace*{1em}}%
  \def\Require{\par\textbf{Require:}~}%
  \def\Return{\textbf{return}~}%
  \def\Comment##1{\hfill\textit{\# ##1}}%
  \def\For##1{\State\textbf{for} ##1\textbf{ do}\par\begingroup\leftskip=1.5em}%
  \def\EndFor{\par\endgroup\State\textbf{end for}}%
}{\par\endgroup}
}
\usepackage{bm}
\usepackage{multirow}
\usepackage{wrapfig}
\usepackage{tikz}
\usepackage{float}
\usepackage{placeins}
\usepackage{caption}

\definecolor{OrthrosPurple}{HTML}{5B2A86}    
\definecolor{OrthrosLink}{HTML}{9D4EDD}      
\definecolor{OrthrosLavender}{HTML}{FAF8FD}  
\definecolor{TikTokPink}{HTML}{5B2A86}
\definecolor{LinkPurple}{HTML}{9D4EDD}
\definecolor{CalGoldHex}{HTML}{FAF8FD}
\hypersetup{allcolors=OrthrosLink}

\newlength{\headlogoh}

\tcbuselibrary{skins}
\newlength{\logoh}
\newlength{\polyuh}
\tcbset{
    titlebox/.style={
        enhanced,
        colback=CalGoldHex,
        colframe=CalGoldHex,
        boxrule=2.5mm,
        arc=2mm,
        auto outer arc,
        left=5mm,
        right=5mm,
        top=5mm,
        bottom=20mm,   
        overlay={%
            \node[anchor=south west, inner sep=0pt]
                at ([xshift=10mm,yshift=5mm]interior.south west)
                {\includegraphics[height=\logoh]{figures/mindlab.pdf}};
            \node[anchor=south east, inner sep=0pt]
                at ([xshift=-10mm,yshift=3.65mm]interior.south east)
                {\includegraphics[height=\polyuh]{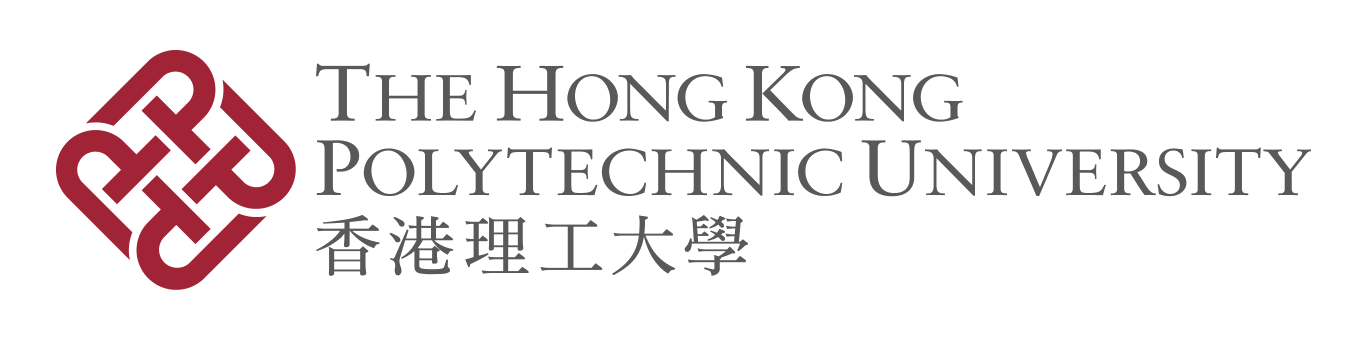}};
        }
    }
}

\IfFileExists{XCharter.sty}{}{\microtypesetup{expansion=false}}
\title{Orthros: Phase-Aware Heterogeneous Attention for Efficient Transformers}
\runningtitle{Orthros: Phase-Aware Heterogeneous Attention for Efficient Transformers}
\author{
Yuhong Chou$^{\dagger}$, Zehao Liu$^{\dagger}$,
Yuqi Pan$^{\ddagger}$, Qian Liu$^{\S}$,
Xianwei Chen$^{\dagger}$, Zhongzhu Zhou$^{\diamond}$,
Jibin Wu$^{\dagger}$\\ 
\begin{tabular}{cc}
$^{\dagger}$The Hong Kong Polytechnic University
&
$^{\ddagger}$University of Chinese Academy of Sciences\\
$^{\S}$Independent Researcher
&
$^{\diamond}$University of Sydney
\end{tabular}
}
\correspondingauthor{Yuhong Chou: yuhong.chou@connect.polyu.hk; Corresponding author: Jibin Wu}

\begin{document}
\begin{abstract}

Efficient attention mechanisms, such as linear and sliding window attention, aim to reduce the computational complexity of Transformers from quadratic to linear. However, their degraded in-context recall often necessitates interleaving full attention layers, leaving the quadratic bottleneck unresolved in long-context processing. To address this, we introduce Orthros attention, which shifts the efficiency paradigm from structural allocation to phase-aware decoupling. As a unified module, it dynamically switches its computation pattern across different phases of sequence processing, enabling linear complexity prefilling and full attention decoding. Our design grants Orthros attention a lower asymptotic complexity compared to vanilla Transformers in prevalent long-context scenarios characterized by heavy prefilling and light decoding. Beyond theoretical formulation and complexity analysis, extensive empirical evaluations demonstrate Orthros's superiority across multiple dimensions. First, we validate its fundamental language modeling capabilities, demonstrating scalability across varying model sizes and strong competitiveness in comprehensive architectural comparisons. Second, Orthros exhibits exceptional conversion feasibility via a data-efficient Transformer-to-Orthros conversion, yielding performance that even surpasses the original Transformer. Finally, Orthros achieves a $21.9\times$ prefilling speedup over Transformers with end-to-end inference gains, while preserving long-context capabilities, as demonstrated by near-perfect 128K needle-in-a-haystack results and Transformer-level performance on real-world in-context recall tasks. Overall, these advantages in high performance, flexibility, and efficiency position Orthros as a promising candidate to succeed the Transformer in industrial applications.

\vspace{5mm}

\textbf{Date}: October 7, 2026

\textbf{Correspondence}: Jibin Wu~(\href{mailto:jibin.wu@polyu.edu.hk}{jibin.wu@polyu.edu.hk})
\end{abstract}

\maketitle
\section{Introduction}

Efficient attention mechanisms have been extensively investigated to address the growing prevalence of long-context scenarios in modern foundation models~\citep{wang2024beyond}. In applications such as high-resolution multimodal~\citep{bai2025qwen3vl, xu2025qwen3omni}, repository-level code analysis~\citep{cao2026qwen3coder, bogomolov2024long}, and long-horizon agentic workflows~\citep{team2025kimik2, zeng2026glm, zhuang2025self}, interaction sequences routinely span hundreds of thousands to millions of tokens. Under such extreme lengths, the quadratic complexity of vanilla full attention~\citep{vaswani2017attention} becomes a critical bottleneck. While methods like sliding window and linear attention~\citep{dai2019transformer, fu2025sliding, wang2025systematic, shen2024scaling} reduce this complexity to a linear scale, their respective mechanisms of key-value (KV) eviction and state compression severely degrade in-context recall~\citep{arora2024simple, zoology2023}, especially in extended contexts. To mitigate this issue, hybrid Transformers that spatially mix linear and full attention~\citep{li2025minimax, lieber2024jamba, dong2024hymba, team2025kimi, blakeman2025nemotron} have emerged as the prevailing industry solution to balance performance and efficiency. However, such hybridization implies that the model's overall asymptotic complexity remains bottlenecked by its most computationally intensive component: full attention. Consequently, hybrid Transformers ultimately yield only a constant-factor speedup, effectively eclipsing the linear complexity advantage inherent to the efficient attention modules.

We propose Orthros attention, an efficient attention mechanism that maintains a favorable subquadratic complexity property without KV eviction or compression, thereby preserving in-context recall in prevalent long-context scenarios. Its core insight lies in shifting the allocation of computational complexity from the structural dimension to the distinct phases. Specifically, Orthros attention applies sliding window attention across the entire sequence, while selectively performing additional full attention at positions tasked with next-token prediction. By adopting distinct computational paradigms, this mechanism perfectly aligns with modern hardware realities: it maintains highly efficient linear complexity during the compute-intensive prefilling phase, while the heavier decoding computation is effectively masked by inherent memory I/O bottlenecks. Through our algorithmic design, these heterogeneous computations are encapsulated within a unified Orthros attention layer, while simultaneously ensuring strict consistency between training and inference. By functioning as a self-contained module, this formulation does not require broader architectural changes, such as cross-layer attention and decoder-decoder designs, thereby preserving the plug-and-play capability to convert existing well-trained foundation models.

Empirically, we conduct comprehensive experiments to validate Orthros attention across diverse scales and settings. First, using a controlled token budget, we verify the architecture's scalability from 160M to 6.8B parameters and perform detailed component ablations to justify our design choices. Second, we pre-train a 1.3B parameter Orthros model from scratch on 100B tokens, demonstrating that it achieves performance highly comparable to strong Transformer baselines. Third, we validate its plug-and-play capability by converting well-trained base models, specifically Llama3.1-8B and Qwen3-8B. Compared to models continually trained on equivalent data volumes, the converted Orthros models deliver performance comparable to or exceeding the original architecture. Fourth, evaluations on in-context recall tasks show that our pre-trained models match Transformer-level performance; notably, in long-context extension, Orthros achieves near-perfect retrieval accuracy on the 128K needle-in-a-haystack test. Fifth, system-level profiling of prefilling efficiency reveals that Orthros yields up to a $21.9\times$ speedup at an extreme sequence length of one million tokens. Finally, end-to-end inference results confirm that this acceleration translates into tangible overall gains, with the doubled FLOP count during decoding effectively masked by the I/O-bound nature of the phase. Together, our theoretical and empirical results validate that the Orthros Transformer delivers performance and efficiency competitive with or superior to strong Transformer baselines, positioning it as a highly promising successor to the Transformer architecture.

\section{Related Works}
\label{section:related_work}

In this section, we first review efficient attention mechanisms, including linear and sliding window attention. Next, we discuss the decoder-decoder framework, a macro-architecture built upon cross-layer attention.

\subsection{Efficient Attention}

Given an input sequence of token vectors $\{\bm{x}_i\}_{i=1}^T$, where $\bm{x}_i\in\mathbb{R}^{1\times d_{\text{model}}}$ we define the query, key, and value vectors at position $i$ as $\bm{q}_i = \bm{x}_i \bm{W}_Q$, $\bm{k}_i = \bm{x}_i \bm{W}_K$, and $\bm{v}_i = \bm{x}_i \bm{W}_V$, where $\bm{W}_Q, \bm{W}_K, \bm{W}_V \in \mathbb{R}^{d_{\text{model}} \times d}$ are projection matrices. The standard causal self-attention~\cite{vaswani2017attention} output $\bm{o}_t$ at position $t$ is formulated as:
\begin{equation}
\label{equation:self_attention}
    \bm{o}_t = \frac{\sum_{j=1}^{t} \exp\left( \frac{\bm{q}_t \bm{k}_j^\top}{\sqrt{d}} \right) \bm{v}_j}{\sum_{m=1}^{t} \exp\left( \frac{\bm{q}_t \bm{k}_m^\top}{\sqrt{d}} \right)}.
\end{equation}

Since computing the output at each position $t$ requires $t$ inner products, evaluating the entire sequence from $t=1$ to $T$ incurs a quadratic computational complexity of $\mathcal{O}(T^2)$. Consequently, the community has actively pursued efficient attention implementations to mitigate this bottleneck. 

By decomposing the exponential similarity via a kernel feature map $\phi(\cdot)$, standard linear attention~\citep{katharopoulos2020transformers} computes the output recursively using a hidden state matrix $\bm{S}_t$ and a normalization vector $\bm{z}_t$:
\begin{equation}
    \bm{o}_t = \frac{\phi(\bm{q}_t) \bm{S}_t}{\phi(\bm{q}_t) \bm{z}_t}, \quad \text{where} \quad \bm{S}_t = \sum_{j=1}^{t} \phi(\bm{k}_j)^\top \bm{v}_j, \quad \bm{z}_t = \sum_{j=1}^{t} \phi(\bm{k}_j)^\top.
\end{equation}

Following this paradigm, numerous linear attention variants have emerged \citep{gu2023mamba, peng2023rwkv, sun2023retentive, qin2024hgrn2, yang2024parallelizing, chou2024metala, dao2024transformers,chou2025zeco, yang2024gated}. Despite introducing diverse mechanisms for hidden state updates and output computations, these models universally preserve the desirable $\mathcal{O}(T)$ asymptotic complexity.

Alternatively, directly restricting the receptive field of self-attention to a local window of size $w$ yields sliding window attention~\citep{2023arXiv231006825J, openai2025gptoss120bgptoss20bmodel, singh2026arcee, ren2024samba}:
\begin{equation}
    \bm{o}_t = \frac{\sum_{j=\max(1, t-w+1)}^{t} \exp\left( \frac{\bm{q}_t \bm{k}_j^\top}{\sqrt{d}} \right) \bm{v}_j}{\sum_{m=\max(1, t-w+1)}^{t} \exp\left( \frac{\bm{q}_t \bm{k}_m^\top}{\sqrt{d}} \right)}.
\end{equation}

For both linear and sliding window attention, the computational overhead at each time step $t$ is bounded by a constant state size or a fixed window size $w$. Therefore, processing the full sequence achieves an overall linear complexity of $\mathcal{O}(T)$.

\subsection{Cross-Layer Attention and Decoder-Decoder Architectures}
In contrast to self-attention, cross-attention~\citep{vaswani2017attention, brandon2024reducing, wu-etal-2025-systematic} derives its queries and key-value (KV) pairs from entirely distinct token sequences. Formally, given a query token $\bm{x}_i$ and a context token $\bm{y}_j$, the projections become $\bm{q}_i = \bm{x}_i \bm{W}_Q$, $\bm{k}_j = \bm{y}_j \bm{W}_K$, and $\bm{v}_j = \bm{y}_j \bm{W}_V$, with the output computed via Equation~\ref{equation:self_attention}. In modern LLMs, this mechanism is frequently used to share KV pairs across different layers, known as cross-layer attention. 

Building upon cross-layer attention, the decoder-decoder framework~\citep{sun2024you, lin2025reconstructing, ren2025decoder, wu2026yoco++} is a specific hybrid macro-architecture that partitions the network into two stages. The first half, termed the \textit{self-decoder}, applies efficient attention and generates a single, shared KV cache. The second half, the \textit{cross-decoder}, relies entirely on full cross-layer attention to attend to this shared cache. In this way, the cross-decoder computation can be entirely skipped before token generation, while the global receptive field is maintained during the decoding phase.

While this architecture achieves a notable balance between efficiency and performance, it inherently possesses two critical limitations. First, relying on a single shared KV cache for half of the network constitutes an extreme sharing strategy, which existing studies on cross-layer attention suggest inevitably compromises performance. Second, this macro-design strictly limits architectural flexibility. Specifically, inserting any state-maintaining attention mechanisms (such as standard self-attention or linear attention) into the cross-decoder would fundamentally break its core efficiency property: the early exit capability before generation.

\section{Method}
\label{section:method}

\begin{figure}[!t]
    \centering
    \includegraphics[width=\textwidth]{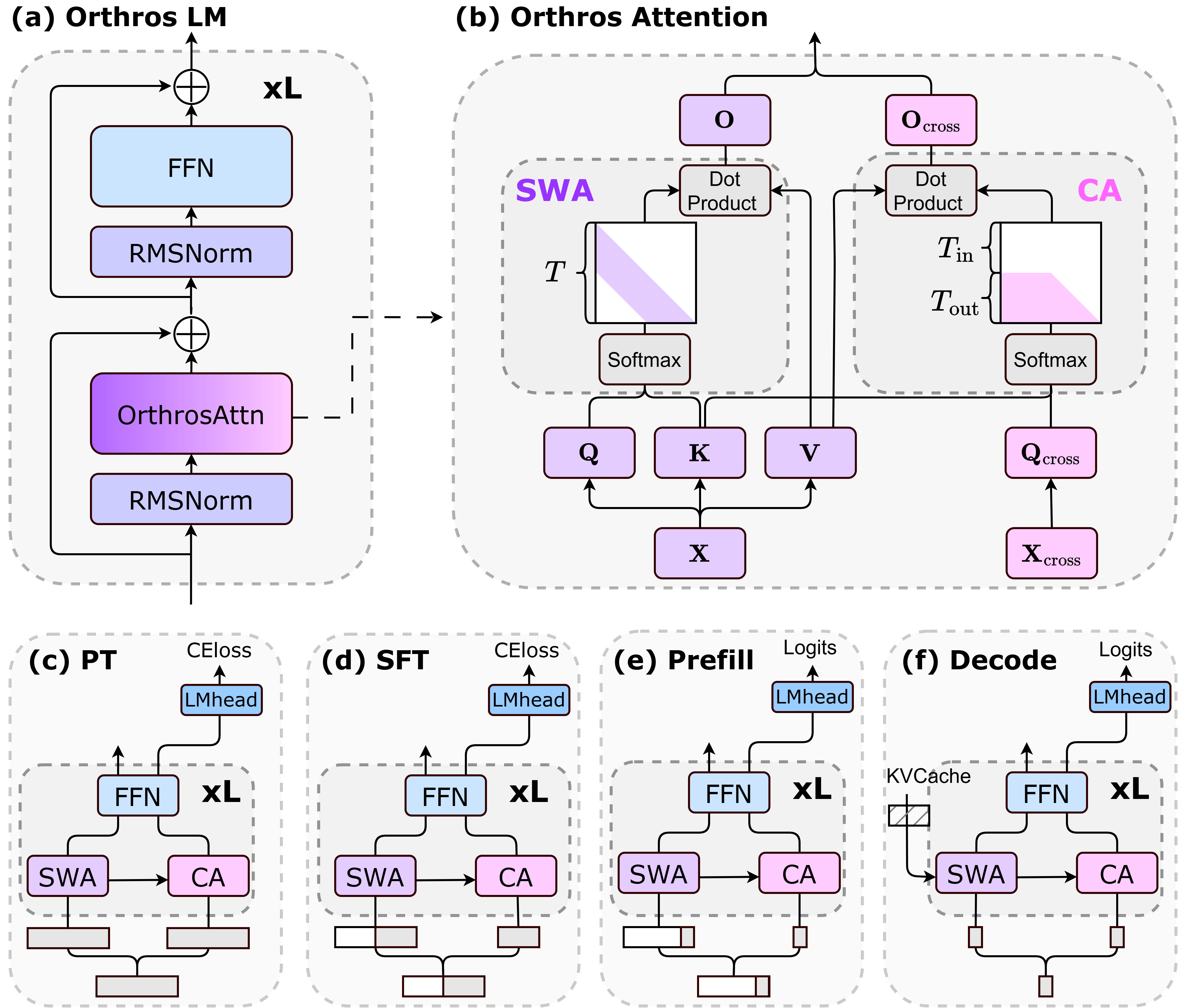}
    \caption{
    Overview of Orthros attention. The top two sub-figures describe the Orthros architecture and the bottom four sub-figures show the comparison of four sequence-processing regimes. SWA stands for sliding window attention, and CA stands for full cross-attention.
    }
    \label{figure:main_figure}
\end{figure}

First, Section~\ref{section:Orthros_algorithm} introduces Orthros attention, a prefilling-decoding heterogeneous attention mechanism that processes input and output sequences using distinct computation paradigms. Next, Section~\ref{section:training_inference_scheme} details the training and inference schemes of the Orthros Transformer from a macro-architectural perspective. Finally, Section~\ref{subsection:complexity_analysis} analyzes the efficiency characteristics of the proposed design across various scenarios.

\subsection{Orthros attention}
\label{section:Orthros_algorithm}

Orthros attention is fundamentally designed around the inherent structure of autoregressive generation, where a sequence comprises a given prompt and subsequent generated tokens. Exploiting this structure, Orthros attention decouples context assimilation from output generation. Specifically, it applies sliding window attention across all tokens to efficiently assimilate local context and materialize a layer-wise KV cache. For positions tasked with output generation, it additionally performs full cross-attention against this materialized KV cache to retrieve global information. 

This mechanism naturally exhibits prefilling-decoding heterogeneity across inference stages: during the prefilling phase, the model predominantly executes the highly efficient sliding window attention, invoking full cross-attention exclusively at the final prompt token to initiate generation; conversely, during the decoding phase, each generation step processes a decoupled pair of tokens: one executes sliding window attention to update the KV cache, while the other performs full cross-attention to predict the subsequent output.

Formally, let $T$ denote the total length of a tokenized sequence (excluding the \texttt{<eos>} token), which serves as the complete model input for next-token prediction. We partition this sequence into two functional segments. Let $T_{\text{out}}$ denote the number of positions tasked with predicting the generated output tokens (spanning from the final prompt token to the last generated token predicting \texttt{<eos>}). Consequently, we define $T_{\text{in}} = T - T_{\text{out}}$ as the length of the preceding sequence. Given a sequence of base token vectors $\{\bm{x}_i\}_{i=1}^T$ and an additional set of output-specific token vectors $\{\bm{x}'_i\}_{i=T_{\text{in}}+1}^T$ maintained exclusively for the $T_{\text{out}}$ output positions, we first compute the key and value vectors for all $i \in [1, T]$ via standard linear projections: $\bm{k}_i = \bm{x}_i \bm{W}_K$ and $\bm{v}_i = \bm{x}_i \bm{W}_V$. For the queries, we apply distinct affine transformations to differentiate their functional roles. Specifically, the query for the base token is computed for all $i \in [1, T]$ as $\bm{q}_i = \bm{\alpha} \odot (\bm{x}_i \bm{W}_Q) + \bm{\beta}$. In contrast, the query for the output-specific token is derived exclusively for the output sequence $i \in [T_{\text{in}}+1, T]$ as $\bm{q}'_i = \bm{\alpha}' \odot (\bm{x}'_i \bm{W}_Q) + \bm{\beta}'$. Here, $\bm{W}_Q, \bm{W}_K, \bm{W}_V \in \mathbb{R}^{d_{\text{model}} \times d}$ denote the standard projection matrices, while $\bm{\alpha}, \bm{\beta}, \bm{\alpha}', \bm{\beta}' \in \mathbb{R}^{1 \times d}$ represent learnable transformation vectors, with $\odot$ denoting element-wise multiplication.

Orthros attention computes outputs via two distinct streams. The sliding window attention output $\bm{o}_t$ (with window size $w$) is computed for all $t \in [1, T]$, while the full cross-attention output $\bm{o}'_t$ is computed exclusively for the output sequence $t \in [T_{\text{in}}+1, T]$:

\noindent
\begin{minipage}{0.48\linewidth}
\begin{equation}
\label{equation:sliding_window_attention}
    \bm{o}_t = \frac{\sum_{j=\max(1, t-w+1)}^{t} \exp\left( \frac{\bm{q}_t \bm{k}_j^\top}{\sqrt{d}} \right) \bm{v}_j}{\sum_{m=\max(1, t-w+1)}^{t} \exp\left( \frac{\bm{q}_t \bm{k}_m^\top}{\sqrt{d}} \right)}
\end{equation}
\end{minipage}%
\hfill
\begin{minipage}{0.48\linewidth}
\begin{equation}
\label{equation:full_cross_attention}
    \bm{o}'_t = \frac{\sum_{j=1}^{t} \exp\left( \frac{\bm{q}'_t \bm{k}_j^\top}{\sqrt{d}} \right) \bm{v}_j}{\sum_{m=1}^{t} \exp\left( \frac{\bm{q}'_t \bm{k}_m^\top}{\sqrt{d}} \right)}
\end{equation}
\end{minipage}

Algorithm~\ref{algorithm:Orthros_attention} outlines the forward pass of the Orthros attention layer, with highlighted lines denoting deviations from standard attention. For computational convenience, the input tensor concatenates the $T$ base tokens and the $T_{\text{out}}$ output-specific tokens along the sequence dimension. The first $T$ tokens generate the shared KV cache and the base queries $Q$ for sliding window attention. Conversely, the subsequent $T_{\text{out}}$ tokens exclusively generate the global queries $Q_{\text{cross}}$ for full cross-attention. The resulting outputs from both streams are concatenated along the sequence dimension before the final linear projection. A visual representation of this architecture is provided in Figure~\ref{figure:main_figure}(b).

\begin{algorithm}[h]
\small
\caption{OrthrosAttention\_Layer}
\label{algorithm:Orthros_attention}
\begin{algorithmic}[1]
\Require Input tensor $X \in \mathbb{R}^{(T + T_{\text{out}}) \times d_{\text{model}}}$, number of heads $H$, sliding window size $W$
\State \textbf{Parameters:} $W_Q, W_K, W_V, W_O \in \mathbb{R}^{d_{\text{model}} \times d_{\text{model}}}$
\State $X, \textcolor{TikTokPink}{X_{\text{cross}}} \gets X_{:T, :}, X_{T:, :}$ \Comment{Split base and output-specific tokens}
\State $K, V \gets X W_K, X W_V$ \Comment{Generate shared KV cache}
\State $Q, \textcolor{TikTokPink}{Q_{\text{cross}}} \gets \bm{\alpha} \odot (X W_Q) + \bm{\beta}, \textcolor{TikTokPink}{\bm{\alpha}' \odot (X_{\text{cross}} W_Q) + \bm{\beta}'}$ \Comment{Distinct query transformations}
\State \textbf{Split heads:}
\Statex \hspace{\algorithmicindent}$Q, K, V \in \mathbb{R}^{T \times H \times d_h}$, $\textcolor{TikTokPink}{Q_{\text{cross}}} \in \mathbb{R}^{T_{\text{out}} \times H \times d_h}$ ($d_h = d_{\text{model}}/H$)
\State $O \gets \text{Flash-Attn}(Q, K, V, \text{window\_size}=\textcolor{TikTokPink}{W})$
\State \textcolor{TikTokPink}{$O_{\text{cross}} \gets \text{Flash-Attn}(Q_{\text{cross}}, K, V, \text{window\_size}=\infty)$}
\State \textbf{Merge heads:} $O \in \mathbb{R}^{T \times d_{\text{model}}}$, $\textcolor{TikTokPink}{O_{\text{cross}}} \in \mathbb{R}^{T_{\text{out}} \times d_{\text{model}}}$
\State \textcolor{TikTokPink}{$O \gets \text{Concat}[O, O_{\text{cross}}] \in \mathbb{R}^{(T + T_{\text{out}}) \times d_{\text{model}}}$}
\State $Y \gets O W_O$
\State \Return $Y$
\end{algorithmic}
\end{algorithm}

At the macro-architectural level, integrating these output-specific tokens requires minimal modifications to the standard Transformer backbone. Non-attention modules like Feed-Forward Networks (FFNs) and Normalization operate strictly token-wise. Consequently, they naturally process the additional tokens alongside the base tokens without requiring any structural changes. As detailed in Algorithm~\ref{algorithm:Orthros_transformer}, the Orthros backbone simply duplicates the embeddings of the $T_{\text{out}}$ output positions and appends them to the sequence. After uniformly processing this concatenated sequence through $L$ layers, the model extracts exclusively these $T_{\text{out}}$ output-specific tokens to compute the final logits.

\begin{algorithm}[h]
\small
\caption{Orthros Transformer Backbone}
\label{algorithm:Orthros_transformer}
\begin{algorithmic}[1]
\Require Input tensor $X \in \mathbb{R}^{T \times d_{\text{model}}}$, total sequence length $T$, output sequence length $T_{\text{out}}$, number of layers $L$
\State \textcolor{TikTokPink}{$Z \gets \text{Concat}[X, X_{T-T_{\text{out}}:, :}] \in \mathbb{R}^{(T + T_{\text{out}}) \times d_{\text{model}}}$} \Comment{Duplicate the $T_{\text{out}}$ output-specific tokens}
\For{$\ell = 1$ to $L$}
    \State $A \gets \text{OrthrosAttention\_Layer}_{\ell}(\text{Attn\_Norm}_{\ell}(Z))$
    \State $Z \gets Z + A$
    \State $F \gets \text{FFN\_Layer}_{\ell}(\text{FFN\_Norm}_{\ell}(Z))$
    \State $Z \gets Z + F$
\EndFor
\State \textcolor{TikTokPink}{$Z \gets Z_{T:, :}$} \Comment{Extract the $T_{\text{out}}$ output-specific tokens}
\State $Y \gets \text{Norm}(Z)$
\State \Return $Y$
\end{algorithmic}
\end{algorithm}

\subsection{Training \& Inference Schemes}
\label{section:training_inference_scheme}

The architectural elegance of the Orthros Transformer lies in its unified formulation, which ensures strict algorithmic consistency between training and inference. We categorize its operational pipelines into four distinct paradigms: Pre-training (PT), Supervised Fine-Tuning (SFT), Prefilling, and Decoding. Remarkably, all four scenarios are seamlessly instantiated without any structural modifications, simply by adjusting the context length $T_{\text{in}}$ and the output length $T_{\text{out}}$. Figure~\ref{figure:main_figure}(c)-(f) provides a visual representation of these configurations.

\textbf{Pre-Training} ($T_{\text{in}} = 0, T_{\text{out}} = T$): In standard causal language modeling, every token serves as a target for next-token prediction. Consequently, the entire sequence is duplicated ($T_{\text{out}} = T$) to instantiate the output-specific tokens, ensuring that full cross-attention is computed at every position to derive the training loss.

\textbf{Supervised Fine-Tuning} ($T_{\text{in}} = T - T_{\text{out}}, T_{\text{out}} \ge 1$): During SFT, the loss is computed exclusively over the generated responses, while the user prompts serve purely as context. Thus, $T_{\text{out}}$ corresponds exactly to the total length of the target responses. Notably, alignment tuning paradigms, such as reinforcement learning on post-rollout data, share this identical formulation, as optimization is similarly restricted to the generated tokens.

\textbf{Prefilling} ($T_{\text{in}} = T - 1, T_{\text{out}} = 1$): When processing a given prompt during inference, all tokens except the last act strictly as contextual inputs ($T_{\text{in}} = T - 1$). The model only duplicates the final prompt token ($T_{\text{out}} = 1$) to perform full cross-attention and initiate the generation of the first output token.

\textbf{Decoding} ($T_{\text{in}} = 0, T_{\text{out}} = 1$): In the autoregressive decoding phase, each step processes exactly one generated token ($T=1$). This single token is duplicated to form the decoupled pair introduced in Section~\ref{section:Orthros_algorithm}: the base token updates the KV cache and computes sliding window attention, while the output-specific token performs full cross-attention against the historical cache to predict the subsequent token.

\subsection{Complexity Analysis}
\label{subsection:complexity_analysis}

We analyze the computational complexity of the Orthros Transformer with respect to the total sequence length $T = T_{\text{in}} + T_{\text{out}}$, model width $d_{\text{model}}$, and sliding window size $W$. 

The model's computation comprises three primary components: sliding window attention ($\mathcal{O}(T W d_{\text{model}})$), full cross-attention ($\mathcal{O}(T_{\text{out}} T d_{\text{model}})$), and token-wise linear projections ($\mathcal{O}((T + T_{\text{out}}) d_{\text{model}}^2)$). In practice, since the window size $W$ is typically small ($W \ll d_{\text{model}}$), the sliding window attention term is asymptotically dominated by the linear projections and can be safely omitted from the bottleneck analysis.

In prevalent long-context scenarios, the output length $T_{\text{out}}$ is typically a lower-order term relative to $T$ ($T_{\text{out}} \ll T$), or even bounded by a constant (e.g., structured tool invocations in agentic workflows). When $T_{\text{out}} \ll T$, the overall complexity is dominated by the cross-attention term $\mathcal{O}(T_{\text{out}} T d_{\text{model}})$, achieving a highly favorable subquadratic scaling. If $T_{\text{out}} \approx \mathcal{O}(1)$, the attention overhead becomes negligible, reducing the model to a strictly linear complexity $\mathcal{O}(T d_{\text{model}}^2)$ governed by the projections. Only in the extreme edge case where $T_{\text{out}} \approx \mathcal{O}(T)$ (e.g., pre-training) does the complexity degrade to the $\mathcal{O}(T^2 d_{\text{model}})$ scaling of a vanilla Transformer.

Finally, during the memory I/O-bound decoding phase, Orthros processes a decoupled pair of tokens per generation step. Fortunately, this additional computation incurs negligible latency overhead. Since the I/O complexity is strictly dominated by loading the weight matrices, the arithmetic cost of processing two tokens, or even a small batch of vectors, is entirely masked by the memory access latency. Furthermore, in batched decoding scenarios, this strategy of increasing active compute tokens enables the system to more rapidly transcend the I/O bottleneck, effectively escaping the regime of inefficient GPU utilization.

\section{Experiments}
\label{section:experiments}

We conduct comprehensive experiments to evaluate Orthros from five perspectives: 
First, we do scaling and ablation studies from 160M to 6.8B parameters to validate the scalability and the key component design.
Second, pre-training from scratch shows that Orthros achieves competitive or superior performance compared with Transformer and efficient attention architectures on benchmarks.
Third, conversion experiments on Llama3.1 and Qwen3 base models demonstrate that existing full attention Transformers can be converted into Orthros with data-efficient continual training.
Fourth, in-context recall evaluations show that Orthros achieves competitive performance among all architectures and near-perfect accuracy on the 128K needle-in-a-haystack test.
Fifth, prefilling efficiency tests in a single GPU and multi GPUs show that Orthros substantially reduces long-context prefilling latency, achieving up to a $21.9\times$ speedup. 
Finally, end-to-end inference evaluations reveal that this prefilling advantage translates into tangible overall speedups. Moreover, the doubled FLOP count during decoding is effectively masked by the I/O-bound nature of this phase, strongly corroborating our theoretical predictions.

In all the experiments, Orthros uses sliding window attention of 128 window size. 
We report all the experiment settings in Appendix~\ref{append:experiment_settings}.

\subsection{Scaling and Ablation Study}
\label{section:scaling_study}

\begin{figure}[htbp]
    \centering
    \includegraphics[width=0.8\textwidth]{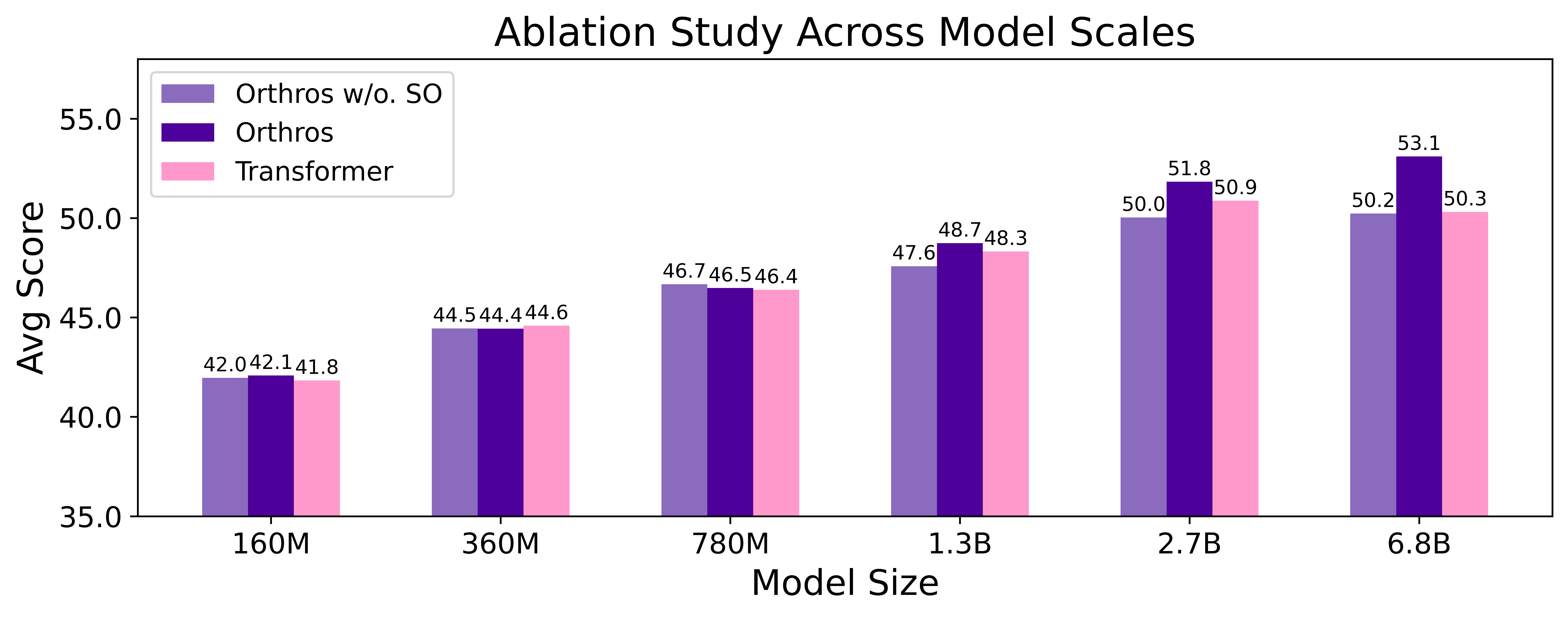}
    \caption{
    The ablation and scaling study of Orthros. The transformer serves as a baseline. Orthros w/o SO removes different scalers $\bm{\alpha}$ and offsets $\bm{\beta}$.
    }
    \label{fig:Orthros_ablation}
\end{figure}

In this section. We conduct scaling and architecture component ablation studies. A key design choice in Orthros is the scale-offset (SO)~\citep{hua2022transformer}, which specializes the shared query projection for different attention streams through lightweight element-wise transformations.
This design preserves parameter sharing and I/O efficiency.

In the experiment, all ablation models are trained on 10B tokens with a sequence length of 2048 and a 0.25M tokens batch size. Orthros denotes our model and its w/o SO variant removes the learnable scale-offset transformation. We evaluate three models of different model sizes from 160M to 6.8B with \texttt{lm-evaluation-harness} framework~\citep{eval-harness} on Common Sense Reasoning benchmarks, which include ARC-challenge (ARC-C)~\citep{allenai:arc}, ARC-easy (ARC-E)~\citep{allenai:arc}, HellaSwag (HS)~\citep{zellers2019hellaswag}, OpenBookQA (OBQA)~\citep{OpenBookQA2018}, PIQA~\citep{Bisk2020}, Social-IQa (SIQA)~\citep{sap2019socialIQa}, WinoGrande (Wino)~\citep{levesque2012winograd}. We report the average score in Figure~\ref{fig:Orthros_ablation}, and full results are reported in Appendix~\ref{append:experiment_metrics}.
These results show that Orthros exhibits competitive model scalability against Transformer baselines, with its scale-offset design providing a better trade-off between parameter efficiency and performance

\subsection{Training from Scratch}
\label{section:traning_from_scratch}

\begin{table}[htbp]
\centering
\caption{Common Sense Reasoning evaluation of different architectures at 1.3B scale. Higher is better for all metrics.}
\label{tab:arch_common_sense_1_3b}
\begin{tabular}{l|ccccccc|c}
\toprule
\textbf{Model}
& \textbf{ARC-C}
& \textbf{ARC-E}
& \textbf{HS}
& \textbf{OBQA}
& \textbf{PIQA}
& \textbf{SIQA}
& \textbf{Wino}
& \textbf{Avg.} \\
& Norm
& 
& Norm
& Norm
& 
& 
& 
& \\
\midrule
Transformer & 37.80 & 70.58 & 56.21 & 37.80 & \textbf{72.91} & 41.10 & \textbf{57.46} & 53.41 \\
Orthros      & \textbf{39.08} & \textbf{71.09} & \textbf{57.19} & \textbf{40.20} & 71.82 & \textbf{42.12} & 56.35 & \textbf{53.98} \\
GDN         & 38.82 & 70.66 & 56.28 & \textbf{40.20} & 71.44 & 40.63 & 55.01 & 53.29 \\
GLA         & 36.86 & 67.97 & 53.29 & 38.20 & 70.73 & 40.07 & 56.20 & 51.90 \\
RetNet      & 35.84 & 69.82 & 54.33 & 37.00 & 71.93 & 39.76 & 55.01 & 51.96 \\
SWA         & 36.95 & 70.16 & 56.11 & 38.80 & 72.42 & 40.94 & 57.14 & 53.22 \\
\bottomrule
\end{tabular}
\end{table}

Next, we pre-train Orthros from scratch and compare it with Transformer and efficient attention architectures, including GatedDeltaNet (GDN), Gated Linear Attention (GLA), RetNet, and sliding window attention (SWA). 
For the 1.3B-scale comparison, all models are trained on 100B tokens fineweb-edu~\citep{lozhkov2024fineweb-edu} dataset with a sequence length of 2048 and a 2M tokens batch size. 
The SWA baseline uses a window size of 512, while Orthros uses a window size of 128, following our default setting. 

We evaluate the models on Common Sense Reasoning benchmarks (Section~\ref{section:scaling_study}), as shown in Table~\ref{tab:arch_common_sense_1_3b}.
Orthros achieves the best average performance among all compared architectures while maintaining comparable performance across individual tasks. 
This result indicates that Orthros preserves strong general language modeling ability. 

\subsection{Conversion Training}
\label{section:conversion}
\begin{table}[!htbp]
\centering
\small
\caption{Evaluation results of converting Llama3.1 and Qwen3 to Orthros. Higher is better for all metrics.}
\label{tab:conversion}
\begingroup
\setlength{\tabcolsep}{3.5pt}
\begin{tabular}{ll|cccccccc|c}
\toprule
\textbf{Base} & \textbf{Model}
& \textbf{ARC-C}
& \textbf{ARC-E}
& \textbf{HS}
& \textbf{OBQA}
& \textbf{PIQA}
& \textbf{SIQA}
& \textbf{Wino}
& \textbf{MMLU}
& \textbf{Avg.} \\
& 
& Norm
& 
& Norm
& Norm
& 
& 
& 
& 
& \\
\midrule
\multirow{3}{*}{Llama3.1}
& Base              & 53.50 & 81.57 & 78.99 & 45.20 & 79.76 & 46.67 & 73.09 & 65.00 & 65.47 \\
& Orthros Conversion & 57.08 & \textbf{84.13} & \textbf{79.10} & \textbf{45.60} & \textbf{80.25} & \textbf{48.93} & \textbf{74.35} & 68.65 & \textbf{67.26} \\
& Base CPT          & \textbf{57.51} & 83.80 & 78.93 & 45.40 & \textbf{80.25} & 47.80 & 73.40 & \textbf{68.89} & 67.00 \\
\midrule
\multirow{3}{*}{Qwen3}
& Base              & 57.34 & 81.73 & \textbf{78.49} & 42.20 & \textbf{79.27} & \textbf{54.61} & \textbf{72.53} & 76.82 & 67.87 \\
& Orthros Conversion & \textbf{59.39} & \textbf{84.81} & 78.06 & \textbf{44.80} & \textbf{79.27} & 50.51 & 72.38 & 77.76 & \textbf{68.37} \\
& Base CPT          & 57.59 & 83.29 & 77.59 & 43.20 & 78.78 & 50.00 & 71.98 & \textbf{77.96} & 67.55 \\
\bottomrule
\end{tabular}
\endgroup
\end{table}

We next study whether existing pretrained full attention models can be converted into Orthros with data-efficient continual pretraining.
Specifically, we convert Llama3.1-8B~\citep{grattafiori2024llama} and Qwen3-8B~\citep{yang2025qwen3} base models to Orthros by replacing their original attention modules with Orthros attention, and continue pre-training both the converted and base models on an 80B-token corpus of high-quality Common Crawl (CC) data~\citep{su2025nemotron} with a restricted mixture of math~\citep{mahabadi2025nemotron} and code~\citep{huang2025opencoder}.
To isolate the effect of architecture conversion from the effect of additional training data, we include two baselines for each model family:
The first is the original base models (Base), which measure the performance before continual training.
The second is Base CPT, which continues training the original full attention architecture on the same 80B-token budget.

We evaluate the models on Massive Multitask Language Understanding (MMLU) and Common Sense Reasoning benchmarks (Section~\ref{section:scaling_study}). The results are shown in Table~\ref{tab:conversion}.
Across both Llama3.1-8B and Qwen3-8B, Orthros Conversion consistently outperforms both the Base and the same-architecture Base CPT baseline on average, indicating gains of Orthros conversion come from the architecture itself rather than merely from additional training.
Additionally, these results demonstrate that Orthros can be obtained not only by pre-training from scratch, but also by converting strong pretrained full attention Transformer with a moderate continual training cost.

\subsection{In-Context Recall Capacity}
\label{section:recall}

\begin{wraptable}{r}{0.5\linewidth}
\vspace{-2mm}
\centering
\small
\caption{In-context recall-intensive task evaluation at 1.3B scale. Higher is better for all metrics. }
\label{tab:arch_recall_intensive_1_3b}
\begingroup
\setlength{\tabcolsep}{3.0pt}
\begin{tabular}{l|cccccc|c}
\toprule
\textbf{Model}
& \textbf{DP}
& \textbf{NQ}
& \textbf{SW}
& \textbf{TQ}
& \textbf{FD}
& \textbf{SQ}
& \textbf{Avg.} \\
\midrule
Trans. & 23.2 & \textbf{22.4} & 36.7 & \textbf{62.3} & 16.4 & \textbf{42.6} & 33.9 \\
Orthros & \textbf{23.3} & 20.8 & \textbf{39.9} & 59.8 & \textbf{24.1} & 38.7 & \textbf{34.4} \\
GDN    & 22.8 & 20.6 & 21.8 & 58.2 & 12.0 & 35.1 & 28.4 \\
GLA    & 21.4 & 16.7 & 21.4 & 58.6 & 10.7 & 33.6 & 27.1 \\
RetNet & 21.0 & 15.8 & 17.6 & 57.7 & 8.3 & 33.7 & 25.7 \\
SWA    & 22.8 & 14.1 & 17.3 & 61.0 & 16.3 & 41.1 & 28.8 \\
\bottomrule
\end{tabular}
\endgroup
\vspace{-4mm}
\end{wraptable}
To evaluate whether Orthros avoids the inherent disadvantage of efficient architectures on recall-intensive tasks, 
we first compare the 1.3B Orthros models (trained from scratch in Section~\ref{section:traning_from_scratch}) on a suite of in-context recall-intensive benchmarks, which include DROP~\citep{Dua2019DROP}, NarrativeQA (NQ)~\citep{narrativeqa}, SWDE (SW)~\citep{arora2023language}, TriviaQA (TQ)~\citep{2017arXivtriviaqa}, FDA (FD)~\citep{arora2023language}, SQuAD (SQ)~\citep{rajpurkar-etal-2016-squad}, as shown in Table~\ref{tab:arch_recall_intensive_1_3b}.
The results show that Orthros achieves the best average performance among all architectures and is comparable with the full attention Transformer on each task.
This suggests that Orthros substantially avoids the inherent recall gap of efficient models, while retaining the efficiency benefits.

\begin{figure}[htbp]
    \centering
    \includegraphics[width=0.8\textwidth]{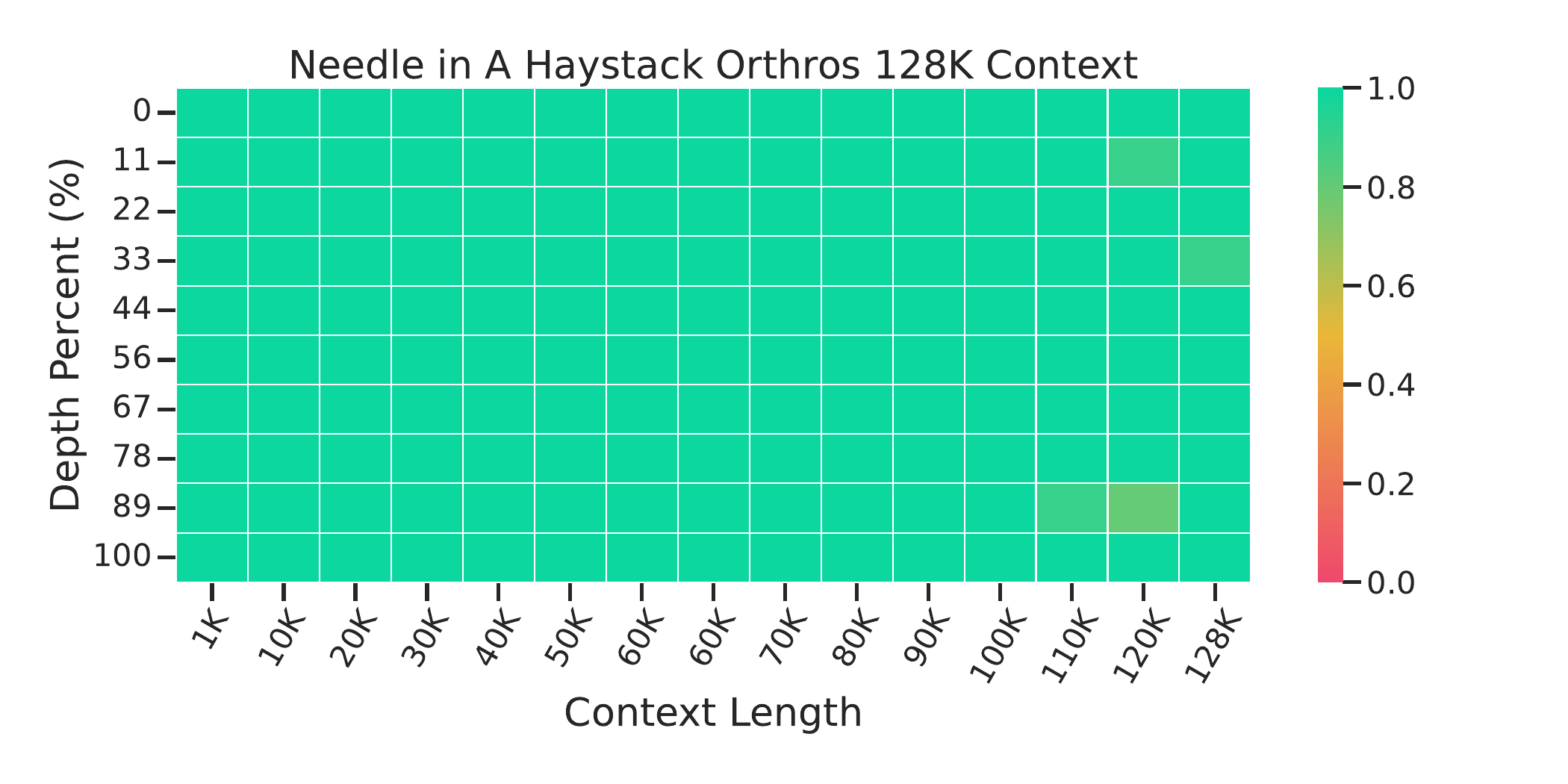}
    \caption{
    Orthros 128K Needle in a Haystack task result. Orthros achieves near-perfect accuracy.
    }
    \label{fig:Orthros_niah}
\end{figure}

We further evaluate the long-range retrieval ability of Orthros on the needle-in-a-haystack benchmark~\citep{kamradt2023needle}.
For this experiment, we use the 8B Orthros model converted from Qwen3 (Section ~\ref{section:conversion}) and further perform 20B-token long-context extension training on ProLong data~\citep{gao2025prolong}.
As shown in Figure~\ref{fig:Orthros_niah}, Orthros maintains near-perfect retrieval accuracy at a context length of 128K.
This result confirms that Orthros preserves strong in-context retrieval ability after long-context extension.

\subsection{Prefilling Efficiency}
\label{section:prefill}
\begin{figure}[!htbp]
    \centering
    \includegraphics[width=\textwidth]{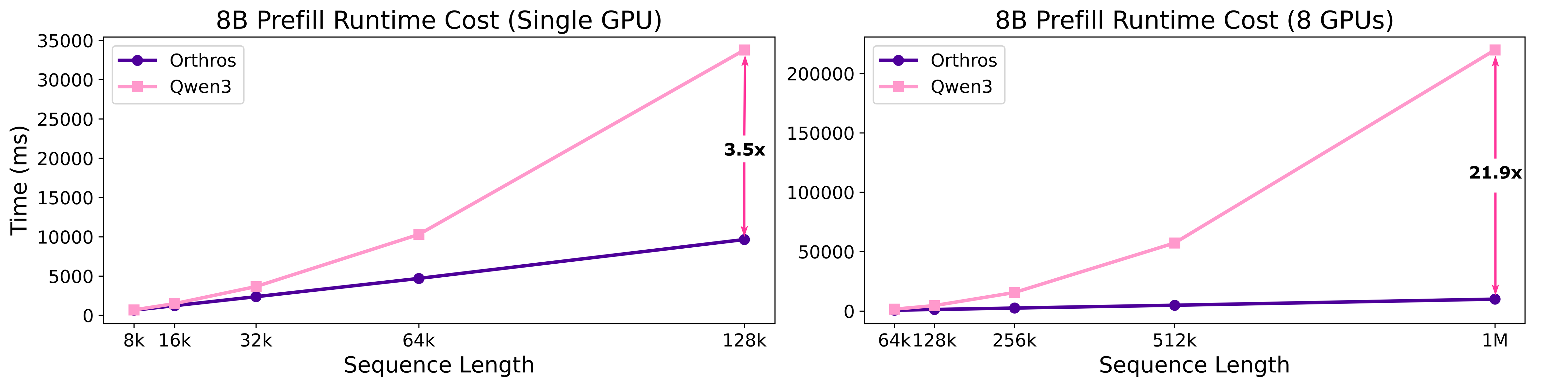}
    \caption{The left figure shows the time cost of Orthros and Qwen3 with a single GPU for prefilling, while the right figure presents 8-GPU prefilling time.}
    \label{fig:Orthros_prefill}
\end{figure}

We benchmark the prefilling efficiency of 8B-Orthros and compare it with Qwen3~\citep{yang2025qwen3}.
Specifically, our prefilling benchmark is implemented using FLA-style~\citep{yang2024fla} models with high-performance kernels.
As shown in Figure~\ref{fig:Orthros_prefill}, Orthros consistently achieves lower prefilling latency on a single GPU for input lengths from 8K to 128K and 8 GPUs for input lengths from 64K to 1M.
The advantage becomes more pronounced as the sequence length increases, which is consistent with the linear-complexity prefilling design of Orthros.

\FloatBarrier

\subsection{End-to-End Inference}
\label{sec:e2e_inference}

We conduct end-to-end inference experiments on a single H100 GPU using
the same SGLang inference engine. The Transformer and Linear Hybrid
baselines follow the 8B architectures of Qwen3 and Qwen3.5, respectively.
Orthros is built on the Qwen3-8B architecture using Orthros attention.
Orthros-H extends Orthros with a layer-wise mixture of full-attention
and linear-attention layers at a ratio of 1:3.
In Orthros-H, the GDN layer, GDN as cross-attention part in which
the query bypasses the convolution module. 
All experiments use a batch size of 2, with input and output sequence lengths of 128, 1K, and 8K tokens, utilizing the Triton backend for attention.

\begin{table}[h]
\centering
\caption{End-to-end inference on a single H100 GPU using SGLang. Time to first token (TTFT) and end-to-end latency (E2E) are measured in seconds; output and decode-only
throughput are measured in tokens per second. Output throughput is
the total number of generated tokens divided by E2E latency.
Lower latency and higher throughput are better.}
\label{tab:e2e_inference}
\scriptsize
\setlength{\tabcolsep}{2pt}
\renewcommand{\arraystretch}{1.15}
\begin{tabular}{@{}l*{3}{cccc}@{}}
\toprule
& \multicolumn{4}{c}{128K input / 128 output}
& \multicolumn{4}{c}{127K input / 1K output}
& \multicolumn{4}{c}{120K input / 8K output} \\
\cmidrule(lr){2-5}
\cmidrule(lr){6-9}
\cmidrule(l){10-13}
Model
& \shortstack{TTFT\\(s)}
& \shortstack{E2E\\(s)}
& \shortstack{Output\\(tok/s)}
& \shortstack{Decode-only\\(tok/s)}
& \shortstack{TTFT\\(s)}
& \shortstack{E2E\\(s)}
& \shortstack{Output\\(tok/s)}
& \shortstack{Decode-only\\(tok/s)}
& \shortstack{TTFT\\(s)}
& \shortstack{E2E\\(s)}
& \shortstack{Output\\(tok/s)}
& \shortstack{Decode-only\\(tok/s)} \\
\midrule
Transformer
& 77.249 & 84.911 & 3.0 & 33.7
& 76.160 & 137.715 & 14.9 & 33.2
& 67.770 & 548.357 & 29.9 & 34.1 \\
\textbf{Orthros}
& \textbf{17.016} & \textbf{24.647} & \textbf{10.4} & 33.3
& \textbf{17.301} & \textbf{76.345} & \textbf{26.8} & 33.7
& \textbf{15.210} & \textbf{488.830} & \textbf{33.5} & 34.6 \\
Linear Hybrid
& 25.566 & 28.130 & 9.1 & 98.8
& 25.173 & 46.308 & 44.2 & 96.8
& 22.747 & 189.302 & 86.5 & 98.3 \\
\textbf{Orthros-H}
& \textbf{10.184} & \textbf{12.771} & \textbf{20.0} & 97.3
& \textbf{9.882} & \textbf{30.724} & \textbf{66.7} & 105.8
& \textbf{9.302} & \textbf{173.887} & \textbf{94.2} & 99.5 \\
\bottomrule
\end{tabular}
\end{table}
\FloatBarrier

\paragraph{Prefilling and decoding.}
As shown in Table~\ref{tab:e2e_inference}, Orthros and Orthros-H
substantially reduce prefilling latency relative to Transformer and
Linear Hybrid, respectively. Their decode-only throughput remains
comparable to that of the corresponding baselines.
The two hybrid models decode faster than the two full-attention
decoding models.

\paragraph{End-to-end latency.}
Faster prefilling translates into lower end-to-end (E2E) latency
and higher output throughput, particularly when the output is short.
As the output length increases, decoding occupies a larger share
of the total time, reducing the relative E2E advantage.

\section{Discussion, Limitation and Future Work}

\paragraph{Discussion.} 
In this work, we introduced Orthros attention, a novel efficient attention mechanism defined by its inherent prefilling-decoding heterogeneity. Extensive empirical evaluations demonstrate that Orthros delivers substantial gains in inference efficiency without compromising downstream performance or long-context capabilities, standing as a compelling alternative to strong Transformer baselines. Beyond these empirical successes, the introduction of prefilling-decoding heterogeneity invites a broader discussion on the fundamental trade-offs in architecture design and the evolving computational paradigms of modern LLMs:
\begin{itemize}
    \item \textbf{Training--Inference Trade-off.}
    Orthros achieves substantial inference gains at the cost of doubling token computation during training to maintain strict training--inference consistency. However, conventional scaling analyses primarily assess architectural efficiency through the relationship between training FLOPs and loss, leaving inference efficiency outside the evaluation. This perspective can therefore undervalue architectures that trade additional training computation for sustained inference savings.

    \item \textbf{Prefill-Dominated Workloads and Compute Allocation.}
    Prefill-heavy applications, such as multimodal understanding and agentic LLM workflows, are particularly well suited to Orthros. Its heterogeneous design allocates less computation to processing inputs and more to generating outputs, reflecting an intuitive strategy for preserving output quality. However, how to fairly compare this redistribution of computation with the uniform computation of standard Transformers remains an open question.
\end{itemize}
Together, these considerations motivate a broader evaluation framework for heterogeneous architectures that accounts for training cost, phase-specific inference cost, and model quality across diverse workloads. Such a framework should make comparison criteria explicit and assess how architectural benefits vary with input and output lengths and deployment scale, enabling a more comprehensive assessment of computational efficiency.

\paragraph{Limitation and Future Work.} 
While Orthros attention demonstrates significant potential, it currently requires computing a doubled sequence during training. In addition, the memory efficiency of its KV cache remains unaddressed. Beyond these limitations, several future efforts must be undertaken to translate this potential into reality for industrial applications. First, Mixture-of-Experts scenarios require specialized designs to prevent decoupled token pairs from being routed to different experts during decoding, which would otherwise exacerbate the I/O bound. Second, the agentic reinforcement learning paradigm for Orthros needs to be thoroughly investigated; this will help demonstrate its efficiency advantages in multi-turn and long-context agentic applications. Finally, adapting Orthros to modern inference frameworks requires further exploration, particularly regarding how to optimize its decoding mechanism across diverse scenarios, such as single-sample generation and batched decoding.

\clearpage
\section*{Acknowledgments}
This work was partially supported by the Research Grants Council of the Hong Kong SAR (Grant No. PolyU25216423, PolyU15217424), and The Hong Kong Polytechnic University (P0056503, P0053699).

We are grateful to Yin Zheng and Zejun Ma for their insightful discussions and support.

\section*{Contributors}

\textbf{Yuhong Chou} led the project, proposed the core idea, designed and conducted the experiments, wrote and revised the paper, and coordinated the author team throughout.

\textbf{Zehao Liu} conducted part of the experimental evaluation, prepared the figures, and contributed to the writing of the paper.

\textbf{Yuqi Pan} developed the initial implementation of the inference framework and contributed substantively to the technical discussions throughout the project.

\textbf{Qian Liu} advised on the formulation of the core idea and on the writing of the paper.

\textbf{Xianwei Chen} contributed to the exploratory experiments on model training.

\textbf{Zhongzhu Zhou} advised on the optimization of the inference framework and on the design of the end-to-end inference experiments.

\textbf{Jibin Wu} supervised the project, secured the funding and resources that supported this work, and is responsible for all correspondence regarding this work.

\clearpage
\bibliography{references}
\appendix
\clearpage
\section{Experiment Settings}
\label{append:experiment_settings}
\begin{table}[!htbp]
\small
\centering
\caption{
Model configurations and training settings for the scaling and ablation study.
All models are trained with a sequence length of 2048 for approximately 10B tokens.
Orthros uses a sliding window size of 128.
}
\label{tab:scaling_ablation_details}
\small
\begin{tabular}{lcccccc}
\toprule
\textbf{Item} 
& \textbf{160M} 
& \textbf{360M} 
& \textbf{780M} 
& \textbf{1.3B} 
& \textbf{2.7B} 
& \textbf{6.8B} \\
\midrule
\multicolumn{7}{c}{\textbf{Model}} \\
\midrule
Architecture 
& \multicolumn{6}{c}{Transformer / Orthros / Orthros w/o SO} \\
Hidden size 
& 768 & 1024 & 1536 & 2048 & 2560 & 4096 \\
Layers 
& 16 & 24 & 24 & 24 & 32 & 32 \\
Heads 
& 6 & 8 & 12 & 16 & 20 & 32 \\
Head dimension 
& \multicolumn{6}{c}{128} \\
Vocabulary size 
& \multicolumn{6}{c}{32000} \\
Torch dtype 
& \multicolumn{6}{c}{bfloat16} \\
Window size 
& \multicolumn{6}{c}{128 for Orthros and Orthros w/o SO} \\
\midrule
\multicolumn{7}{c}{\textbf{Training}} \\
\midrule
Training tokens 
& \multicolumn{6}{c}{10B} \\
Sequence length 
& \multicolumn{6}{c}{2048} \\
Training steps 
& \multicolumn{6}{c}{40K} \\
Global batch size 
& \multicolumn{6}{c}{0.25M tokens} \\
Seed
& \multicolumn{6}{c}{42} \\
Optimizer 
& \multicolumn{6}{c}{AdamW} \\
Learning rate 
& \multicolumn{6}{c}{$3\times 10^{-4}$} \\
LR scheduler 
& \multicolumn{6}{c}{Cosine} \\
Warmup steps 
& \multicolumn{6}{c}{1500} \\
Minimum LR ratio 
& \multicolumn{6}{c}{0.001} \\
Weight decay 
& \multicolumn{6}{c}{0.05} \\
Gradient clipping 
& \multicolumn{6}{c}{1.0} \\
Loss 
& \multicolumn{6}{c}{Cross Entropy} \\
FSDP 
& \multicolumn{6}{c}{Full shard} \\
Data parallel 
& \multicolumn{6}{c}{8} \\
\bottomrule
\end{tabular}
\end{table}

\begin{table}[!htbp]
\centering
\caption{
Model configurations and continual training settings for Transformer-to-Orthros conversion.
Base denotes the original pretrained checkpoint without additional training.
Orthros Conversion replaces the original full attention modules with Orthros Attention.
Base CPT continues training the original Transformer architecture under the same training setting.
}
\label{tab:conversion_training_details}
\scriptsize
\begin{tabular}{lccc|ccc}
\toprule
\textbf{Item}
& \multicolumn{3}{c|}{\textbf{Llama3.1-8B}}
& \multicolumn{3}{c}{\textbf{Qwen3-8B}} \\
\cmidrule(lr){2-4}
\cmidrule(lr){5-7}
& \textbf{Base}
& \textbf{Orthros Conv.}
& \textbf{Base CPT}
& \textbf{Base}
& \textbf{Orthros Conv.}
& \textbf{Base CPT} \\
\midrule
\multicolumn{7}{c}{\textbf{Model}} \\
\midrule
Base checkpoint
& \multicolumn{3}{c|}{Llama3.1-8B}
& \multicolumn{3}{c}{Qwen3-8B-Base} \\
Architecture
& Transformer & Orthros & Transformer
& Transformer & Orthros & Transformer \\
Hidden size
& \multicolumn{3}{c|}{4096}
& \multicolumn{3}{c}{4096} \\
Layers
& \multicolumn{3}{c|}{32}
& \multicolumn{3}{c}{36} \\
Vocabulary size
& \multicolumn{3}{c|}{128256}
& \multicolumn{3}{c}{151936} \\
Torch dtype
& \multicolumn{3}{c|}{bfloat16}
& \multicolumn{3}{c}{bfloat16} \\
window size
& -- & 128 & --
& -- & 128 & -- \\
\midrule
\multicolumn{7}{c}{\textbf{continual Training}} \\
\midrule
Additional training
& No
& \multicolumn{2}{c|}{80B tokens}
& No
& \multicolumn{2}{c}{80B tokens} \\
Sequence length
& --
& \multicolumn{2}{c|}{32768}
& --
& \multicolumn{2}{c}{32768} \\
Global batch size
& --
& \multicolumn{2}{c|}{8M tokens}
& --
& \multicolumn{2}{c}{8M tokens} \\
Training steps
& --
& \multicolumn{2}{c|}{10000}
& --
& \multicolumn{2}{c}{10000} \\
Seed
& --
& \multicolumn{2}{c|}{42}
& --
& \multicolumn{2}{c}{42} \\
Optimizer
& --
& \multicolumn{2}{c|}{AdamW}
& --
& \multicolumn{2}{c}{AdamW} \\
Learning rate
& --
& \multicolumn{2}{c|}{$3\times10^{-5}$}
& --
& \multicolumn{2}{c}{$3\times10^{-5}$} \\
LR scheduler
& --
& \multicolumn{2}{c|}{Cosine}
& --
& \multicolumn{2}{c}{Cosine} \\
Warmup steps
& --
& \multicolumn{2}{c|}{500}
& --
& \multicolumn{2}{c}{500} \\
Minimum LR ratio
& --
& \multicolumn{2}{c|}{0.1}
& --
& \multicolumn{2}{c}{0.1} \\
Weight decay
& --
& \multicolumn{2}{c|}{0.1}
& --
& \multicolumn{2}{c}{0.1} \\
Gradient clipping
& --
& \multicolumn{2}{c|}{1.0}
& --
& \multicolumn{2}{c}{1.0} \\
FSDP
& --
& \multicolumn{2}{c|}{Full shard}
& --
& \multicolumn{2}{c}{Full shard} \\
Data parallel
& --
& \multicolumn{2}{c|}{32}
& --
& \multicolumn{2}{c}{32} \\
Activation checkpointing
& --
& \multicolumn{2}{c|}{Yes}
& --
& \multicolumn{2}{c}{Yes} \\
\bottomrule
\end{tabular}
\end{table}

In this section, we report all the experiment settings. All the training was conducted in H100 GPUs, 80GB memory \\
In Table~\ref{tab:scaling_ablation_details}, we report all the experiment settings of the ablation and scaling experiments in Section~\ref{section:scaling_study}. Section~\ref{section:traning_from_scratch}, 1.3B Models uses the same configuration as shown in Table~\ref{tab:scaling_ablation_details}.
For efficient models, we use the implementation of FLA~\citep{yang2024fla}.
In Table~\ref{tab:conversion_training_details}, we report all the experiment settings of the ablation and scaling experiments in Section~\ref{section:conversion}. 

In Section~\ref{section:recall}, we use the 8B-Orthros converted from Qwen3-8B for in-context recall-intensive benchmarks and trained on 20B long context for the needle-in-a-haystack task. For Section~\ref{section:prefill}, we use the 8B-Orthros converted from Qwen3-8B, sequence parallelism was enabled in multi GPUs experiment.

\section{Experiment Metrics}
\label{append:experiment_metrics}

We report all ablation experiment results in this section, as showm in Table~\ref{tab:ablation_scale_study}. 

\begin{table}[!htbp]
\centering
\caption{Ablation study across different model scales. We report normalized accuracy for ARC-Challenge, ARC-Easy, HellaSwag, OpenBookQA, and PIQA, and accuracy for WinoGrande.}
\label{tab:ablation_scale_study}
\resizebox{\linewidth}{!}{
\begin{tabular}{llccccccc}
\toprule
\textbf{Param} & \textbf{Model} 
& \textbf{ARC-C/Norm} 
& \textbf{ARC-E/Norm} 
& \textbf{HS/Norm} 
& \textbf{OBQA/Norm} 
& \textbf{PIQA/Norm} 
& \textbf{Wino} 
& \textbf{Avg.} \\
\midrule
\multirow{3}{*}{160M}
& Orthros w/o SO          & 25.86 & \textbf{47.64} & 33.14 & \textbf{32.20} & 62.62 & 50.28 & 41.96 \\
& Orthros w/ SO       & \textbf{26.19} & 46.68 & \textbf{34.26} & 30.80 & \textbf{63.33} & 51.22 & \textbf{42.08} \\
& Transformer     & 25.77 & 46.25 & 33.64 & 30.60 & 62.08 & \textbf{52.64} & 41.83 \\
\midrule
\multirow{3}{*}{360M}
& Orthros w/o SO          & \textbf{28.58} & \textbf{51.68} & 38.01 & 32.20 & 64.74 & 51.46 & 44.45 \\
& Orthros w/ SO       & 27.90 & 50.25 & \textbf{38.70} & 32.80 & 65.23 & \textbf{51.78} & 44.44 \\
& Transformer     & 28.41 & 50.00 & 38.52 & \textbf{33.80} & \textbf{65.94} & 50.83 & \textbf{44.58} \\
\midrule
\multirow{3}{*}{780M}
& Orthros w/o SO         & 29.35 & 53.07 & 41.86 & \textbf{35.00} & 67.14 & \textbf{53.59} & \textbf{46.67} \\
& Orthros w/ SO       & \textbf{29.78} & 53.41 & \textbf{43.32} & 33.40 & 67.57 & 51.38 & 46.48 \\
& Transformer     & 29.44 & \textbf{53.75} & 42.52 & 33.40 & \textbf{67.95} & 51.30 & 46.39 \\
\midrule
\multirow{3}{*}{1.3B}
& Orthros w/o SO         & 30.03 & 55.05 & 44.83 & 34.40 & \textbf{69.31} & 51.85 & 47.58 \\
& Orthros w/ SO       & \textbf{31.48} & 56.19 & \textbf{46.64} & \textbf{34.80} & 69.04 & \textbf{54.30} & \textbf{48.74} \\
& Transformer     & 31.06 & \textbf{57.07} & 45.30 & 33.60 & 68.99 & 53.91 & 48.32 \\
\midrule
\multirow{3}{*}{2.7B}
& Orthros w/o SO         & 33.11 & 58.21 & 48.85 & 35.80 & 69.59 & 54.62 & 50.03 \\
& Orthros w/ SO       & \textbf{34.22} & \textbf{62.92} & \textbf{50.22} & \textbf{36.80} & 70.84 & \textbf{55.96} & \textbf{51.83} \\
& Transformer     & 33.19 & 59.89 & 48.95 & 35.60 & \textbf{71.71} & 55.88 & 50.87 \\
\midrule
\multirow{3}{*}{6.8B}
& Orthros w/o SO         & 32.59 & 60.44 & 49.39 & 36.00 & 70.02 & 52.96 & 50.23 \\
& Orthros w/ SO       & \textbf{35.84} & \textbf{65.57} & \textbf{53.43} & \textbf{36.60} & \textbf{72.69} & \textbf{54.46} & \textbf{53.10} \\
& Transformer     & 33.45 & 60.90 & 49.84 & 34.80 & 70.35 & 52.49 & 50.31 \\
\bottomrule
\end{tabular}
}
\end{table}

\end{document}